# Multiscale Hybrid Non-local Means Filtering Using Modified Similarity Measure

Zahid Hussain Shamsi and Dai-Gyoung Kim*

*Abstract*—**A new multiscale implementation of non-local means filtering for image denoising is proposed. The proposed algorithm also introduces a modification of similarity measure for patch comparison. The standard Euclidean norm is replaced by weighted Euclidean norm for patch based comparison. Assuming the patch as an oriented surface, notion of normal vector patch is being associated with each patch. The inner product of these normal vector patches is then used in weighted Euclidean distance of photometric patches as the weight factor. The algorithm involves two steps: The first step is multiscale implementation of an accelerated non-local means filtering in the stationary wavelet domain to obtain a refined version of the noisy patches for later comparison. This step is inspired by a preselection phase of finding similar patches in various non-local means approaches. The next step is to apply the modified non-local means filtering to the noisy image using the reference patches obtained in the first step. These refined patches contain less noise, and consequently the computation of normal vectors and partial derivatives is more accurate. Experimental results indicate equivalent or better performance of proposed algorithm as compared to various state of the art algorithms.**

## I. INTRODUCTION

THE phenomenon of image degradation is quite natural due to the digitization and quantization of original image. Although image denoising techniques have been extensively studied and effectively employed, the preservation of texture, edges and fine details during denoising the image, is an open problem and needs rigorous treatment.

Prior to implementation of the non-local principle, a variety of variational, PDE-based, wavelet transform-based, wavelet based and local neighborhood filtering methods were proposed relying on the local features for image denoising. A major shift in this direction was initiated by the bilateral filtering [1] which exploits spatial and intensity domain for image densoising. In this approach spatially proximate pixels are given more weights in similarity measure. The wavelet-based BLS-GSM [2] method has provided the best results in terms of PSNR measure; however, these denoised images contain ringing artifacts and have low visual quality. More recently, Baudes *et al*. [3] introduced non-local means filtering for image densoising. Although the PSNR of non-local means filtering was found to be less than the wavelet-based BLS-GSM [2], the notion of patch-based approach combined with the idea of non-locality led to an entirely new way of attacking the problem. Earlier, approaches similar to non-local means were used for image inpainting [4] and texture synthesis [5]. Since the non-local means was proposed, extensive research for better estimation of parameters or finding suitable similarity measure has improved the performance of nonlocal means filtering for a variety of noise models in various image processing applications.

Kervrann *et al*. [6] provided a theoretical foundation for intuitive non-local means approach using Bayesian statistics. In this approach refined adaptive dictionaries of similar patches are obtained using Bayesian estimation,

This work was supported by the National Research Foundation of Korea (NRF) funded by the Korean (MEST) (NRF-2011-0026245).
The authors are with the Department of Applied Mathematics, Hanyang University , Ansan, 426-791, South Korea (email: zshamsi14@hanyang.ac.kr; dgkim@hanyang.ac.kr ).

while irrelevant patches are rejected. Then, these learned dictionaries are used for patch-based comparison using a modified similarity measure. Elad and Aharon [7] proposed the learned dictionaries of patches using KSVD algorithm and then employed sparse representation over the trained dictionaries for image denoising. Tasdizen [8] proposed using the similarity of patches in the principal component analysis domain. Extensive research in patch-based denoising has resulted in state-of- the-art algorithms [9], [10]. Gilboa and Osher [11] provided an elegant interpretation of non-local means approach as a generalization of variational and PDE-based formulations. Brox *et al.* [12] proposed a computationally very efficient algorithm for non-local filtering that arranged the data in a cluster tree. This special arrangement further helps in preselection of similar patches. Also, an iterative version of non-local filtering based on the variational framework was suggested in [12]. In [13], a specific Euclidean space of patches was elegantly defined for implementing PDE based diffusion or smoothing processes.

In addition to the photometric property is used for patch-based comparison, patches contain much more information that requires the attention of researchers. With this motivation, we propose the notion of normal vectors patch corresponding to each photometric or pixel valued patch. By employing this definition in the second phase of our algorithm, we achieve remarkably better results than most of the state-of-the-art algorithms in the presence of high noise. Inspired by the special treatment of central patch in [14], we employ a slightly higher weight than in the standard non-local means approach; the value in our approach is associated with the central patch heuristically. This modification has further improved our results.

The rest of the paper is organized as follows. A brief review of non- local means filtering is provided in section II. The new multiscale two-stage algorithm and its implementation are explained in section III. Experimental results are described in section IV, and the conclusions are drawn in section V.

## II. Non-local Means Filtering Techniques

Consider a gray-scale intensity image $u(i)_{i \in \Omega}$ over a rectangular bounded region $\Omega$ in $\mathrm{IR}^2$ with additive white Gaussian noise. Instead of non-local, it is preferable to use the notion of semi-local filtering. It was indicated in [14] that considering whole image to search for similar patches has no major benefit, with the exception of periodic or quasi-periodic images. Also searching the whole image for each pixel is computationally too expensive. Therefore, in the rest of this paper, the term non-local means refers specifically to semi-local filtering. The NL-means filtering is defined by [3]

$$u_{nlm}(i) = \frac{1}{K(i)} \sum_{j \in \Delta i} w(i,j) u(i) \qquad (1)$$

where $u_{nlm}(i)$ is the denoised value at pixel location *I*, $w(i,j)$ is the weight obtained by determining similarity of noisy patches around central pixel *i* and other pixels *j* in the search window $\Delta i$, and $K(i) = \sum_{j \in \Delta i} w(i,j)$ is the normalization factor. The weights are obtained using Gaussian kernel with the weighted $L_2$ norm as follows [3]:

$$w(i,j)_{i \neq j} = \exp\left(-\frac{\|\mathbf{u}(i) - \mathbf{u}(j)\|_{2,\alpha}^2}{h^2}\right) \qquad (2)$$

where $\mathbf{u}(i)$ is the rectangular patch around pixel *i* being processed, and $\mathbf{u}(j)$ is the neighborhood of same size around the other pixels j in the search window $\Delta i$. A Gaussian Kernel $G_\alpha$ of standard deviation is used to consider the spatial proximity of the central patch and the other patches in the search window. The non-local means approach involves three parameters: size of patch $\mathbf{u}(i)$, size of the search window $\Delta i$, and filtering or smoothing parameter *h*. Discussion on the choices of these parameters can be found in [15] and [14].

To increase the computational efficiency of non-local means filtering, Baudes *et al*. [15] proposed an accelerated version of non-local means algorithm that, instead of using the central pixel, replaces the whole central patch around the central pixel with a weighted average of patches around the other pixels in the search window. Mathematically this process is defined as [15]:

$$\mathbf{u}_{nlm}(i) = \frac{1}{K(i)} \sum_{j \in \Delta i} w(i, j)\mathbf{u}(j) \qquad (3)$$

where the similarity measure is the same as that defined in eq. (2). However, the cost for the improved computational efficiency of the accelerated non-local means is a slight decrease in the PSNR value.

III. THE PROPOSED ALGORITHM

In this paper, we adopt a hybrid approach to non-local means filtering. Our scheme is comprised of two stages as shown in Fig. 1. Firstly, we apply the accelerated version in the wavelet transform domain to get a pre-denoised image. Second, we employ a modified version of conventional nonlocal means filtering on the given noisy image. In this step, the weights are computed using pre-denoised image obtained in the first step.

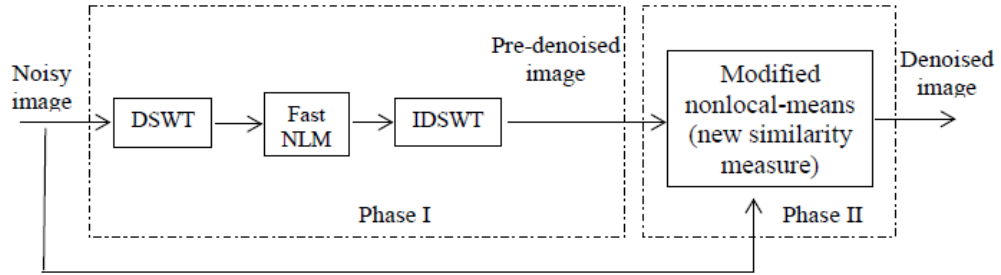

Fig. 1: Schematic representation of the proposed non-local means filtering algorithm

### A. Multiscale Accelerated Non-local Means

Inspired by various approaches [6], [7], [9], based on using pre-denoised images to obtain refined patches, we seek to obtain the pre-denoised images in our approach. However, in contrast to those techniques we adopt a very simple approach. We, first decompose the image $U$ using two dimensional discrete stationary wavelet transform (DSWT) up to the coarsest level J.

$$DSWT_J U = \left(\left(\left(W_{2^j}^x\right), \left(W_{2^j}^y\right), \left(W_{2^j}^{xy}\right)\right)_{1 \le j \le J}, \left(S_{2^j}\right)\right) U \qquad (4)$$

Next, the accelerated version of non-local means filtering [15] is performed on detail bands $\mathbf{W}_{2^j}^x(U)$, $\mathbf{W}_{2^j}^y(U)$ and $\mathbf{W}_{2^j}^{xy}(U)$ for each scale level j=1, 2, …, J. Finally, the reference pre-denoised image is obtained using the inverse stationary wavelet transform. Thanks to an overcomplete, rotationally invariant and sparse representation of a noisy image in the DSWT domain, the image can be efficiently denoised using accelerated nonlocal means filtering without much degradation of texture, edges, and fine details.

### B. Normal Vector Patches and Weight Factor

Assuming the patches as surfaces, we associate normal vectors with the patches, defined as normal vector patch $\mathbf{N}_i$. Notice that the elements of $\mathbf{N}_i$ are normal vectors. We first compute gradient of the whole image and then form the patches of horizontal and vertical components of the gradient vector around each pixel *i*. These patches are denoted by $\mathbf{N}_{i_x}$ and $\mathbf{N}_{i_y}$, where the subscripts $x$ and $y$ represent partial derivatives in horizontal and vertical directions, respectively. The size, $m \times m$, of these patches is the same as the size of the photometric patch. We then compute the element-wise inner product of normal vectors of central patch with the normal vectors of the other patches in the search window $\Delta i$, as follows:

$$\Gamma = \mathbf{N}_i \circledast \mathbf{N}_j = \mathbf{N}_{i_x} \cdot * \mathbf{N}_{j_x} + \mathbf{N}_{i_y} \cdot * \mathbf{N}_{j_y} \tag{5}$$

where $\cdot *$ denotes the point-wise multiplication of patches and $\Gamma$ is the patch or matrix containing all the element-wise inner products. The weight factor is then defined as

$$\eta(i,j)_{i \neq j} = \exp\left(-\frac{1}{(m^2-1)\max_{1 \leq k,l \leq m}\{\Gamma_{kl}\}} \sum_{1 \leq k,l \leq m} \Gamma_{kl}\right) \tag{6}$$

The intuitive motivation for introducing this factor is to acquire the degree of similarity of the original photometric patches, based on the similarity of mean orientation of respective normal vector patches, as shown in Fig. 2. We use this factor in the second step of our algorithm because the computation of derivatives is sensitive to noise level. In the pre-denoised image, the noise level is much lower than in the respective noisy image.

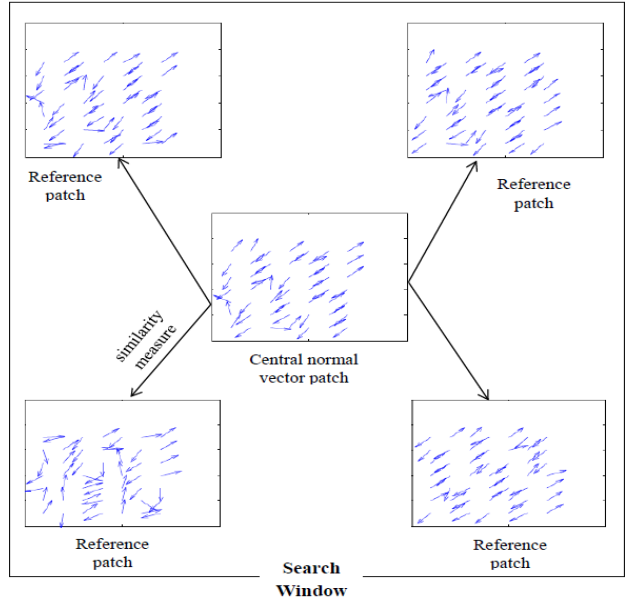

Fig. 2:   Schematic representation of the patch-based comparison of normal vectors

### C. Modified Non-local Means Filtering

After getting pre-denoised image as defined in III-A, we perform the modified version of standard non-local means filtering on given noisy image, based on the similarity of reference patches from the pre-denoised image. The modified similarity measure is defined as

$$w(i,j)_{i \neq j} = \exp\left(-\frac{\|\mathbf{u}(i) - \mathbf{u}(j)\|_2^2 \eta(i,j)}{h^2}\right) \tag{7}$$

It can be noticed that the convolution of the Gaussian kernel $G_\alpha$ in eq. (2) is replaced by the weight factor $\eta(i,j)$ obtained with eq. (6). The convolution with Gaussian kernel $G_\alpha$ was to take into account the spatial proximity of the pixels around central pixel. However, the proposed weight factor is based upon the similarity of the mean normal orientation of central patch with other patches in the search window. Our second modification to standard non-local means filtering is made in the assignment of weight to the central patch when compared with itself. The standard non-local means assigns the maximum of all the computed weights with other patches in the search window [15].

Mathematically, this can be expressed as

$$w(i,i) = \max_{j \in \Delta i; j \neq i} \{w(i, j)\} \quad (8)$$

This choice of weight for self-comparison is intuitive and arbitrary. Salmon [14] discussed the different choices for assigning self-similarity weight. However, we set this weight based on a heuristic approach. For low or medium noise levels in the image, we assign a slightly higher self similarity weight than used in conventional non-local means filtering; our self-similarity weight is defined by

$$w(i,i) = \frac{4}{3} \max_{j \in \Delta i; j \neq i} \{w(i, j)\} \quad (9)$$

The intuitive justification for assigning higher self-similarity weight is based on to the assumption that presence of low or medium levels of noise affects the self similarity slightly. However, for severe noise $(\sigma > 20)$, this assumption is very weak and we use the self similarity weight as defined in eq. (8).

**TABLE I**
PSNR COMPARISON OF THE PROPOSED ALGORITHM WITH VARIOUS STATE-OF-THE-ARTS ALGORITHM

| Image | σ | BM3D | GSM | KSVD | BNLM | Proposed |
|---|---|---|---|---|---|---|
| Lena | 10 | 35.93 | **35.61** | 35.47 | 35.25 | 35.4 |
| | 15 | 34.27 | 33.9 | 33.7 | 33.68 | **33.93** |
| | 20 | 33.05 | 32.66 | 32.38 | 32.63 | **32.75** |
| | 25 | 32.05 | 31.69 | 31.32 | 31.55 | **31.76** |
| | 50 | 28.86 | **28.61** | 27.79 | 27.51 | 28.47 |
| Barbara | 10 | 34.98 | 34.03 | **34.42** | 33.83 | 33.8 |
| | 15 | 33.11 | 31.86 | **32.37** | 32.21 | 32.35 |
| | 20 | 31.78 | 30.32 | 30.83 | 30.88 | **31.13** |
| | 25 | 30.72 | 29.13 | 29.6 | 29.77 | **30.07** |
| | 50 | 27.17 | 25.48 | 25.47 | 24.91 | **26.41** |
| Boats | 10 | 33.92 | 33.58 | **33.64** | 33.18 | 32.94 |
| | 15 | 32.14 | 31.7 | **31.73** | 31.45 | 31.52 |
| | 20 | 30.88 | 30.38 | 30.36 | 30.16 | **30.42** |
| | 25 | 29.91 | 29.37 | 29.28 | 29.11 | **29.43** |
| | 50 | 26.64 | **26.38** | 25.95 | 25.13 | 26.22 |
| Peppers | 10 | 34.68 | 33.77 | **34.28** | 33.87 | 33.53 |
| | 15 | 32.7 | 31.74 | **32.22** | 32.06 | 32.03 |
| | 20 | 31.29 | 30.31 | **30.82** | 30.75 | **30.82** |
| | 25 | 30.16 | 29.21 | 29.73 | 29.77 | **29.78** |
| | 50 | 26.41 | 25.9 | 26.13 | 23.84 | **26.28** |
| House | 10 | 36.71 | 35.35 | **35.98** | 35.67 | 35.78 |
| | 15 | 34.94 | 33.64 | **34.32** | 34.23 | 34.23 |
| | 20 | 33.77 | 32.39 | 33.2 | **33.24** | 33.11 |
| | 25 | 32.86 | 31.4 | 32.15 | **32.3** | 32.15 |
| | 50 | 29.37 | 28.26 | 27.95 | 27.64 | **28.38** |

*D. Summary of the Proposed Algorithm*

The proposed denoising algorithm can be summarized as follows:
1) Construct a scale-space in wavelet domain by decomposing the noisy image using discrete stationary wavelet transform up to the coarsest scale J.
2) Perform the accelerated non-local means filtering [15], as described in III-A to obtain the denoised wavelet

3) Reconstruct the pre-denoised image using inverse DSWT. This image will be used as a reference in the next step for patch based comparison.
4) Denoise the given noisy image using the modified non-local means filtering proposed in III-C. The patch comparison is performed on the refined patches obtained in the previous step.

A schematic diagram of the proposed algorithm is shown in Fig. 1.

## IV. EXPERIMENTAL RESULTS

To evaluate the results of proposed method, benchmark gray-scale images of Lena, Barbara, Boats, Peppers, and House are considered. We use $9 \times 9$ patches, $15 \times 15$ search windows, and $h = 6\sigma$ as the filtering parameter in the first phase. In the second phase, we selected $7 \times 7$ patches with filtering parameter $h = 3.87\sigma$. We can choose the same patch size for both the phases; however, choosing the different patch sizes provides a slightly better result. Also, the filtering parameter is reduced to $3.87\sigma$ from $6\sigma$ to avoid the over-smoothing phenomenon.

**Table II**
PSNR COMPARISON OF AVERAGE PERFORMANCE
FOR DIFFERENT NOISE LEVELS

| σ | BM3D | GSM | KSVD | BNLM | Proposed |
|---|---|---|---|---|---|
| 10 | 35.24 | 34.47 | **34.76** | 34.36 | 34.29 |
| 15 | 33.43 | 32.57 | **32.87** | 32.73 | 32.81 |
| 20 | 32.15 | 31.21 | 31.52 | 31.53 | **31.65** |
| 25 | 31.14 | 30.16 | 30.42 | 30.5 | **30.63** |
| 50 | 27.69 | 26.93 | 26.66 | 25.81 | **27.15** |

The proposed algorithm is compared with various state-of-the-art algorithms: those by Dabov *et al.* [9] (BM3D), Elad and Aharon [7] (KSVD), Portilla *et al.* [2] (BLS-GSM), and Kervrann *et al.* [6] (BNLM). The PSNR value is used as a metric for comparing the denoising capability of each scheme. Table I provides the PSNR comparison values for the above cited approaches; the bold-faced values represent the best performance among the last four columns.

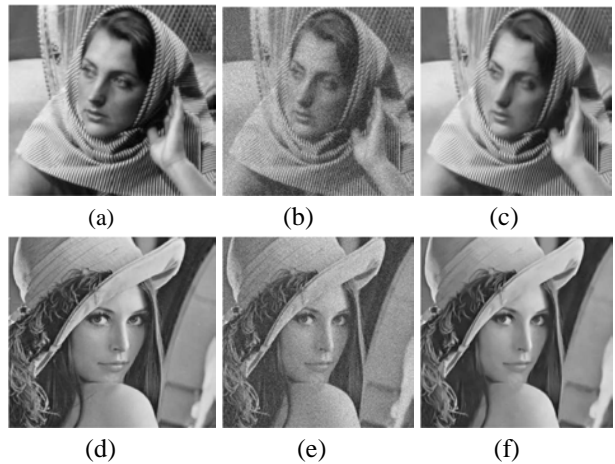

(a)   (b)   (c)
(d)   (e)   (f)

Fig. 3: Left, middle, and right columns represent the original, noisy (σ = 20), and denoised images of Barbara and Lena, respectively.

The results of BM3D are always the best regardless of the noise level. For the rest of the algorithms, K-SVD performance is better when the noise level is low. However, our proposed algorithm yields better results than the others in the presence of moderate and severe noise. In particular, for Barbara image with high texture, our algorithm achieves much better results in terms of PSNR than the others. Our results can also be confirmed based on the average of the performances at various noise levels. Table II indicates that, on average, our algorithm outperforms the other three algorithms in the presence of moderate or severe noise. Fig. 3 shows the visual results for Lena and Barbara images at magnified scale. It can be seen that the texture in both the images is very much preserved, while piece-wise constant regions are not very smoothed out.

## V. Conclusion

A new definition of normal vector patch is proposed to acquire more information about the similarity of photometric patches. The experimental results demonstrate the effectiveness of the proposed algorithm when combined with notion of normal vector patch. Patches continue to offer an undiscovered world of information. Therefore, in future, we plan to further explore their associations with other characteristics, such as coherence and curvature.